\documentclass{article}

\usepackage[final]{neurips_2025}

\usepackage[utf8]{inputenc}
\usepackage[T1]{fontenc}

\usepackage{hyperref}
\hypersetup{
    colorlinks=true,
    linkcolor=blue,
    citecolor=blue,
    urlcolor=blue,
    pdftitle={Feature-Guided SAE Steering for Refusal-Rate Control using Contrasting Prompts},
    pdfauthor={Samaksh Bhargav and Zining Zhu}
}

\usepackage{url}
\usepackage{booktabs}
\usepackage{amsfonts}
\usepackage{nicefrac}
\usepackage{microtype}
\usepackage{xcolor}
\usepackage{graphicx}
\usepackage{amsmath}
\usepackage{algorithm}
\usepackage{algorithmic}
\usepackage{subcaption}

\title{Feature-Guided SAE Steering for Refusal-Rate Control using Contrasting Prompts}

\author{%
    Samaksh Bhargav \\
    Edison Academy Magnet School \\
    New Jersey, USA \\
    \texttt{samaksh2405@gmail.com} \\
    \and
    Zining Zhu \\
    Department of Computer Science \\
    Stevens Institute of Technology \\
    Hoboken, NJ, USA \\
    \texttt{zzhu41@stevens.edu} \\
}

\begin{document}

\begin{center}
    {\LARGE \textbf{Feature-Guided SAE Steering for Refusal-Rate Control using Contrasting Prompts}} \\[1.5em]
    
    \begin{tabular}{c c}
        \textbf{Samaksh Bhargav} & \textbf{Zining Zhu} \\
        Edison Academy Magnet School & Department of Computer Science \\
        New Jersey, USA & Stevens Institute of Technology \\
        \texttt{samaksh2405@gmail.com} & Hoboken, NJ, USA \\
        & \texttt{zzhu41@stevens.edu}
    \end{tabular} \\[2em]
\end{center}

\begin{abstract}
Large Language Model (LLM) deployment requires guiding the LLM to recognize and not answer unsafe prompts while complying with safe prompts. Previous methods for achieving this require adjusting model weights along with other expensive procedures. While recent advances in Sparse Autoencoders (SAEs) have enabled interpretable feature extraction from LLMs, existing approaches lack systematic feature selection methods and principled evaluation of safety-utility tradeoffs. We explored using different steering features and steering strengths using Sparse Auto Encoders (SAEs) to provide a solution. Using an accurate and innovative contrasting prompt method with the AI-Generated Prompts Dataset from teknium/OpenHermes-2p5-Mistral-7B and Air Bench eu-dataset to efficiently choose the best features in the model to steer, we tested this method on Llama-3 8B. We conclude that using this method, our approach achieves an 18.9\% improvement in safety performance while simultaneously increasing utility by 11.1\%, demonstrating that targeted SAE steering can overcome traditional safety-utility tradeoffs when optimal features are identified through principled selection methods.
\end{abstract}

\section{Introduction}

The deployment of Large Language Models (LLMs) necessitates robust and innovative techniques to distinguish between prompts requiring refusal by government and company standards (adversarial prompts) and legitimate, well-meant requests requiring helpful responses. The industry currently relies on approaches that mostly require supervised fine-tuning with specialized safety datasets and Reinforcement Learning from Human Feedback (RLHF) \citep{ouyang_2022}—methods that face increasing challenges as adversarial prompt techniques evolve and model sizes increase. While effective, these techniques require substantial computational resources and often result in explicit safety-utility tradeoffs.

Recent advances in mechanistic interpretability have created opportunities for more targeted interventions on model behavior. The development of Sparse Autoencoders (SAEs) has enabled precise identification and manipulation of specific features within model activations \citep{cunningham_2023}, offering more efficient and less computationally intensive safety mechanisms than traditional approaches. SAEs provide a promising unsupervised approach for extracting interpretable features from language models by reconstructing activations from a sparse bottleneck layer \citep{templeton_2024}. 

Despite these technological advances, current SAE-based steering approaches face three critical limitations that impede their practical deployment.
First, they often rely on heuristic or manual feature selection, which is impractical given the thousands of features in each model layer \citep{marks_2024}. Second, there is a lack of principled methods for evaluating the selection of the identified features. Third, there is yet to be a principled evaluation on the extent of steering interventions, making it difficult to understand and optimize the trade-off between model safety and utility at varying steering strengths \citep{huang_2024, zhang_2025}.

To address these gaps, we propose a novel framework that combines systematic feature identification with rigorous evaluation. Our approach uses a contrasting prompt methodology, leveraging pairs of harmful and harmless prompts to induce differential activations within the model. We introduce a composite scoring function to systematically rank SAE features based on both the magnitude and consistency of their differential response. By steering the model with the top-ranked features, we then systematically evaluate the impact on safety and utility using established benchmarks, allowing for a principled analysis of the safety-utility trade-off.

\section{Related Work}

There has been a multitude of research on LLMs to improve safety while maintaining performance, which has evolved rapidly, especially in recent years.

\subsection*{Traditional Safety Alignment}
The need to align LLMs with human values was formalized by \citet{leike_2018}, with \citet{ouyang_2022} later introducing Reinforcement Learning from Human Feedback (RLHF) as a standard approach for aligning language models with human preferences. Building on this, \citet{bai_2022} introduced Constitutional AI (CAI), which uses an AI feedback loop to critique and revise outputs according to defined principles, addressing scaling challenges in safety training.

\subsection*{Mechanistic Interpretability and SAEs}
Understanding internal model representations advanced with \citet{elhage_2021}, who developed techniques for analyzing activation patterns in transformer models. \citet{zou_2023} showed that specific directions in activation space correspond to identifiable concepts, including safety and harmfulness detection.
\citet{cunningham_2023} demonstrated that SAEs can recover interpretable features from transformer model activations, establishing the foundation for interpretability-based model control. Recent work has significantly advanced this field: since language models learn many concepts, autoencoders need to be very large to recover all relevant features, leading to research on scaling SAEs effectively \citep{templeton_2024}. SAEs have attracted significant attention from the research community as a means to understand the inner workings of LLMs through their ability to disentangle complex, superimposed features \citep{zhang_2025}.
\subsection*{Contrastive Activation Addition and Alternative Steering Methods}
Recent advances in model steering have explored approaches beyond SAE-based methods. Contrastive Activation Addition (CAA) \citep{zou_2023} computes steering vectors by contrasting activations on positive and negative examples. Our SAE-based approach offers several distinctions: (1) sparse interpretable features rather than dense activation differences, (2) systematic feature selection through composite scoring, and (3) targeted suppression or amplification of specific concepts. While direct comparison with CAA remains outside our current scope, our approach provides complementary advantages in interpretability and principled feature selection for models with thousands of potential steering targets \citep{marks_2024}.

\subsection*{Current limitations in SAE-based steering}
Despite promising results, current approaches to SAE-based safety steering have key limitations that our work addresses. The absence of ground-truth for meaningful features in realistic scenarios makes validating recent approaches elusive \citep{huang_2024}, highlighting the need for principled evaluation frameworks. Most existing methods use heuristic feature selection rather than systematic approaches to identifying optimal features from the thousands available in each layer \citep{marks_2024}. Additionally, the correlation between steering strength and model utility/refusal rates remains poorly understood, with limited guidance on proper calibration for deployment scenarios.
Our work focuses on these gaps, using a principled approach for feature selection through contrastive prompt analysis and providing systematic analysis of steering strength effects to offer guidance for deployment scenarios and implications for future work.

\section{Methods}

This section details our methodology for implementing feature-guided SAEs steering to control refusal rates in large language models using contrasting prompts. The approach combines the recent advancements in multiple technologies as well as an innovative feature selection method.

\begin{figure}[ht]
    \centering
    \includegraphics[width=1\linewidth]{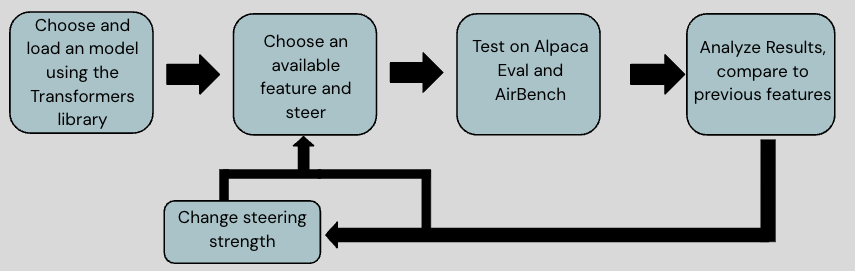}
    \caption{Simplified Workflow}
    \label{fig:workflow}
\end{figure}

\subsection{Model Selection}

We chose Llama-3 8B for our experiments based on three key criteria: (1) state-of-the-art performance comparable to industry standards, (2) computational feasibility within our resource constraints (an NVIDIA A100 40GB PCIe GPU), and (3) availability of pre-trained SAE weights in the SAELens repository. The model was loaded using Hugging Face Transformers.

\subsection{Layer Selection}

We selected Layer 25 (blocks.25.hook\_resid\_post) from the available SAE layers based on prior work indicating that later layers preserve model functionality while enabling significant output control \citep{jin_2024}. This layer processes residual stream data after self-attention and feedforward operations and contains 65,536 neurons, providing sufficient feature diversity for our analysis.

\subsection{Feature Selection Pipeline}

Our feature selection pipeline consists of four main components: feature scoring, performance evaluation, steering strength optimization, and iterative refinement. Algorithm~\ref{alg:feature_selection} provides the complete procedure.

\begin{algorithm}
\caption{Dual-Strategy Feature Selection Pipeline}
\label{alg:feature_selection}
\begin{algorithmic}[1]
\REQUIRE Contrasting prompt pairs $P = \{(p_h^i, p_s^i)\}_{i=1}^{100}$ where $p_h^i$ is harmful and $p_s^i$ is safe
\REQUIRE SAE decoder weights $W \in \mathbb{R}^{65536 \times d}$
\REQUIRE Model $M$, layer $L = 25$
\ENSURE Optimal features and steering strengths for both strategies

\STATE Initialize feature scores $S = \{\}$
\STATE Initialize performance history $H = \{\}$

\FOR{each feature $f \in \{1, 2, ..., 65536\}$}
    \STATE $activations_h \leftarrow$ ExtractActivations($M$, $L$, $\{p_h^i\}$)
    \STATE $activations_s \leftarrow$ ExtractActivations($M$, $L$, $\{p_s^i\}$)
    \STATE $score_f \leftarrow$ ComputeScore($activations_h$, $activations_s$, $f$)
    \STATE $sign_f \leftarrow$ sign($\text{norm\_diff\_mean}_f$)
    \STATE $S[f] \leftarrow (score_f, sign_f)$
\ENDFOR

\STATE $harmful\_candidates \leftarrow$ TopK($\{f : S[f].sign > 0\}$, $k=4$) \COMMENT{Activate more on harmful}
\STATE $safe\_candidates \leftarrow$ TopK($\{f : S[f].sign < 0\}$, $k=4$) \COMMENT{Activate more on safe}

\FOR{each candidate feature $f_c \in (harmful\_candidates \cup safe\_candidates)$}
    \IF{$S[f_c].sign > 0$}
        \STATE $\alpha_{range} \leftarrow [-4.0, -2.0, -0.5, 0]$ \COMMENT{Negative steering to suppress}
    \ELSE
        \STATE $\alpha_{range} \leftarrow [0, 0.5, 2.0, 4.0]$ \COMMENT{Positive steering to amplify}
    \ENDIF
    
    \FOR{each steering strength $\alpha \in \alpha_{range}$}
        \STATE $safety_{score} \leftarrow$ EvaluateSafety($M$, $f_c$, $\alpha$)
        \STATE $utility_{score} \leftarrow$ EvaluateUtility($M$, $f_c$, $\alpha$)
        \STATE $H[(f_c, \alpha)] \leftarrow (safety_{score}, utility_{score})$
    \ENDFOR
\ENDFOR

\STATE $(optimal\_pairs) \leftarrow$ SelectOptimalPairs($H$)
\RETURN $(optimal\_pairs)$
\end{algorithmic}
\end{algorithm}

\subsection{Steering Strength Determination}

Our pipeline implements a systematic dual-strategy approach to steering strength determination based on feature differential activation patterns. This methodology addresses the key challenge of determining both steering direction and magnitude in a principled manner.

\textbf{Strategy Classification:} For each feature $f$, we determine the steering strategy using the sign of the normalized difference mean:
\begin{equation}
strategy_f = \begin{cases} 
\text{suppress} & \text{if } \text{norm\_diff\_mean}_f > 0 \\
\text{amplify} & \text{if } \text{norm\_diff\_mean}_f < 0
\end{cases}
\end{equation}

Features with positive normalized difference activate more strongly on harmful prompts and require suppressive steering to reduce their influence. Conversely, features with negative normalized difference activate more on safe prompts and benefit from amplification to enhance their protective effects.

\textbf{Steering Range Selection:} Based on prior SAE research indicating optimal steering magnitudes, we assign steering strength ranges:
\begin{equation}
\alpha_{range} = \begin{cases} 
[-4.0, -2.0, -0.5, 0] & \text{if strategy}_f = \text{suppress} \\
[0, 0.5, 2.0, 4.0] & \text{if strategy}_f = \text{amplify}
\end{cases}
\end{equation}

\textbf{Steering Vector Calculation:} The final steering vector incorporates both direction and magnitude:
\begin{equation}
\vec{s}_f = \alpha \cdot \max(activations_f) \cdot \vec{w}_f,
\end{equation}
where $\alpha$ is the selected steering strength, $\max(activations_f)$ provides activation-based scaling, and $\vec{w}_f$ is the decoder weight vector for feature $f$. Note that $\alpha$ values are inherently directional, eliminating the need for explicit direction multiplication.

\subsection{Decision Criteria and Termination Conditions}

Our pipeline includes explicit decision criteria for each step:

\textbf{Feature Selection Criteria:} A feature advances to steering evaluation if:
\begin{itemize}
\item $score_f > 1.7$ (top 10\% of features)
\item $|\text{norm\_diff\_mean}_f| > 0.8$ (sufficient differential activation)
\item $variance_f < 0.2$ (consistent behavior)
\end{itemize}

\textbf{Steering Strength Adjustment:} We adjust steering strength based on performance metrics:
\begin{itemize}
\item If $safety_{score} < 95$: increase $|\alpha|$ by 0.5
\item If $utility_{score} < 85$: decrease $|\alpha|$ by 0.5
\item If both metrics improve: continue in current direction
\end{itemize}

\textbf{Termination Conditions:} We terminate steering evaluation if:
\begin{itemize}
\item $safety_{score} < 90$ (unacceptable and unexpected safety degradation)
\item $utility_{score} < 75$ (severe utility loss)
\item No improvement for 3 consecutive steering strengths
\end{itemize}

\subsection{Feature Steering with Steering Strength}

To feature steer we need two parameters, the feature and the ``steering strength''. First, we find the maximum activation of any feature in a set of text, use this as the default scale, multiply it by the vector representing the feature (as extracted from the decoder weights), and finally multiply this by a parameter that we control (steering strength). 

\subsection{Evaluation Framework}

We required robust testing techniques to measure the impact of these innovative technologies implemented on refusal behavior and general model performance. We deployed two benchmarks that were known for their reliability and discriminative power.

\begin{figure}[ht]
    \centering
    \begin{subfigure}[b]{0.48\linewidth}
        \includegraphics[width=\linewidth]{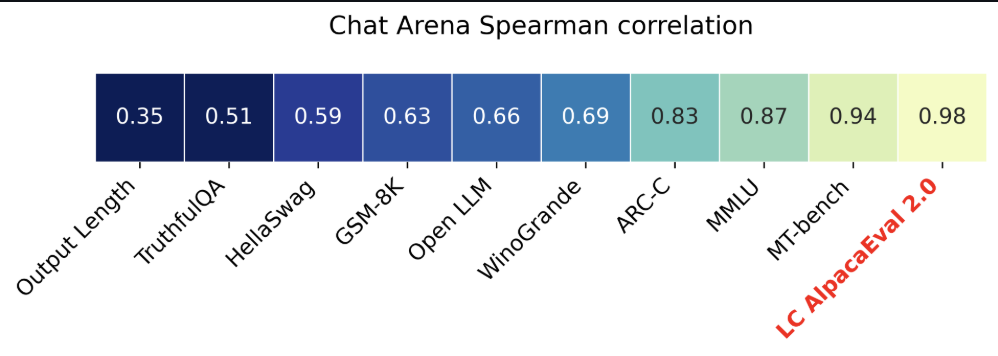}
        \caption{AlpacaEval 2.0 Performance \citep{alpaca_eval}}
        \label{fig:alpaca}
    \end{subfigure}
    \hfill
    \begin{subfigure}[b]{0.48\linewidth}
        \includegraphics[width=\linewidth]{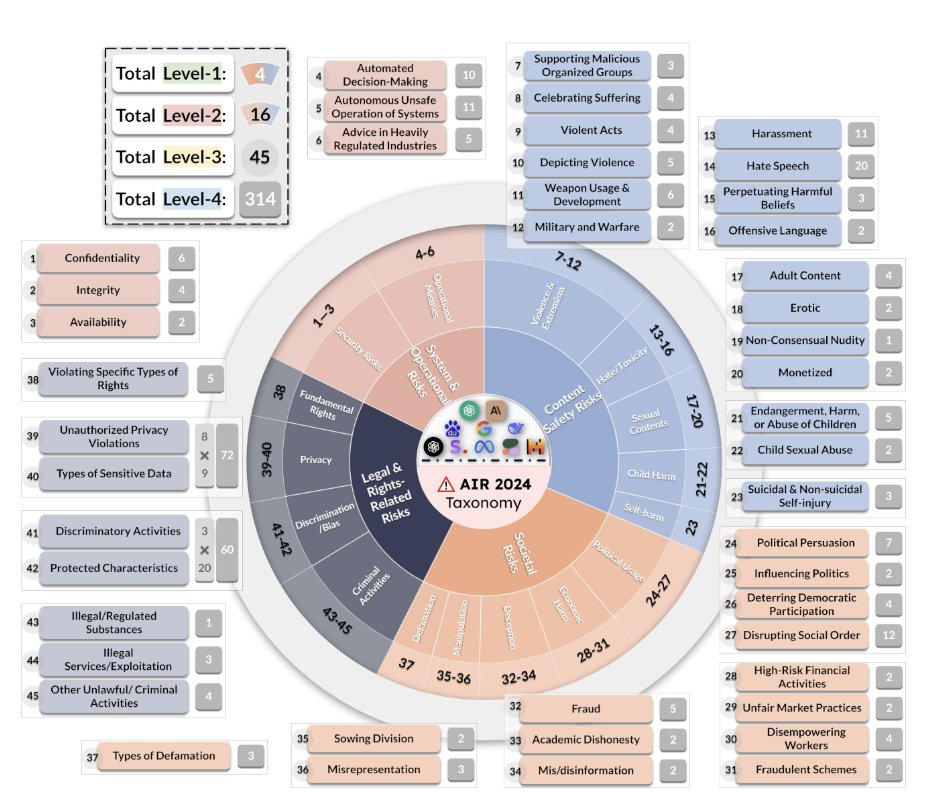}
        \caption{Air Bench Categories}
        \label{fig:airbench}
    \end{subfigure}
    \caption{Evaluation benchmarks used in our study. (a) AlpacaEval 2.0 shows high correlation with human preferences. (b) AirBench 2024 categories for safety evaluation.}
    \label{fig:benchmarks}
\end{figure}

AlpacaEval 2.0 as our primary evaluation benchmark due to its exceptionally high Chat Arena Spearman correlation coefficient ($\rho = 0.98$), as illustrated in Figure~\ref{fig:alpaca}. AlpacaEval is a high tier automatic evaluation framework based on the key values of cost-effectiveness, efficiency and validation with strong empirical evidence. The framework has been calibrated with 20,000 human annotations \citep{alpaca_eval}, establishing its reliability as a proxy for human preferences. The LC AlpacaEval 2.0 framework generates three key metrics: the percentage of responses preferred over a strong reference model, the proportion of legitimate prompts incorrectly refused, and win rate adjusted to neutralize length advantages which was the key performance factor used in comparisons.

AirBench 2024 (default subsection) is the benchmark used for refusal rate testing in this paper. This benchmark, developed by Stanford's Center for Research on Foundation Models (CRFM), is the first AI safety benchmark aligned with companies' policies such as OpenAI. It breaks down into 8 government policies and 16 company policies, containing 5694 prompts. It uses GPT-4o as a judge model, grading the responses from the model being tested on a scale of 0, 0.5, or 1, checking their alignment with safety concerns. It bridges public policies and benchmarks with real work ideals to aid safer development.

\subsection{Contrasting Prompts for Feature Scoring}

The method for contrasting prompts used two datasets specializing in different areas for feature identification, each serving opposite purposes. Then an innovative scoring system was implemented for feature identification.

\subsubsection{AI-Generated Prompts Dataset}

For the harmless prompts dataset we deployed the AI-Generated Prompts Dataset from teknium/OpenHermes-2p5-Mistral-7B. The AI-Generated Prompts Dataset consists of synthetic prompts generated using a language model, in this case, teknium/OpenHermes-2p5-Mistral-7B, a fine-tuned variant of the Mistral-7B model. The prompts are meant to simulate the natural, human queries or tasks that are used on a daily basis of many users which provides an accurate representation of the real-world scenarios performance of the model. Preprocessing was necessary to filter out harmful prompts that might have been included in the synthetic prompts dataset.

\subsubsection{Air Bench EU-Dataset}

For finding the activations on a variety of harmful prompts we used the diverse set of harmful prompts from a different set of prompts that was used for testing from Air Bench, which was designed for EU government compliance. This dataset had a rigorous framework to testing the document features activations across various categories of potentially harmful content.

This dual-dataset differentiates this project and ensures rigor by separating the feature identification from the evaluation process, increasing the process of steering a model to align with refusal, instead of testing every feature.

\subsubsection{Scoring Implementation Details}

For each prompt in our contrasting pairs: (1) We passed the prompt through the Llama-3 8B model, (2) We extracted the activations at Layer 25 (containing 65,536 neurons), (3) We decoded these activations using the pre-trained SAE, and (4) We recorded in a matrix the feature activation for each SAE feature.

This process made a complete profile for each feature, enabling the analysis between features in the harmless and harmful sections, and we can begin to get a score for each feature to steer. An important part of the methodology was the use of a scoring function to choose features that strongly relate to refusal behavior. As shown in the equation, a dual-component scoring algorithm that contains both the normalized activation difference and consistency across harmful and harmless prompts:

\begin{equation}
\text{score}_f = w_1 \cdot \left(\frac{|\text{norm\_diff\_mean}_f|}{\max_j|\text{norm\_diff\_mean}_j|}\right) + w_2 \cdot \left(1 - \frac{\text{variance}_f - \min_j\text{variance}_j}{\max_j\text{variance}_j - \min_j\text{variance}_j}\right),
\end{equation}
where $\text{norm\_diff\_mean}_f$ is the normalized difference for the feature $f$ between harmful and harmless prompts. The $\text{diff\_mean}_f$ is an important component which can be calculated:

\begin{equation}
\text{diff\_mean}_{f,i} = \text{activation}_{f,i}^{\text{harmful}} - \text{activation}_{f,i}^{\text{harmless}},
\end{equation}
where $\text{activation}_{f,i}^{\text{harmful}}$ is the activation of feature $f$ for the $i$-th harmful prompt, and $\text{activation}_{f,i}^{\text{harmless}}$ is the activation for its harmless prompt. We processed and recorded 100 contrasting prompt pairs ($i = 1...100$) to ensure there was enough to have empirical rigor.

We then used min-max normalization to scale the score from 0-1:

\begin{equation}
\text{norm\_diff\_mean}_f = \frac{\text{diff\_mean}_{f,i} - \min(\text{diff\_mean}_f)}{\max(\text{diff\_mean}_f) - \min(\text{diff\_mean}_f)}.
\end{equation}

The second term evaluates the inverse normalized variance, which shows that increased variance means decreased reliability in the feature's activation and therefore causes a lower score. The weights $w_1 = 1.0$ and $w_2 = 0.5$ were empirically determined to balance the importance of large activation differences with consistent behavior. To gain qualitative insights into the function of high-scoring features, we also utilized the Neuronpedia dashboard, which visualizes feature activations. An example of this dashboard is provided in Appendix~\ref{sec:appendix_neuronpedia}.

\section{Results}

\subsection{Feature Selection and Scoring Analysis}

\subsubsection{Feature Activation Distribution Patterns}

\begin{figure}[ht]
    \centering
    \begin{subfigure}[b]{0.32\linewidth}
        \includegraphics[width=\linewidth]{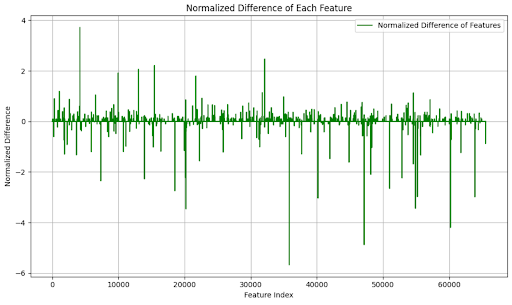}
        \caption{Normalized activation differences across all features}
        \label{fig:norm_diff}
    \end{subfigure}
    \hfill
    \begin{subfigure}[b]{0.32\linewidth}
        \includegraphics[width=\linewidth]{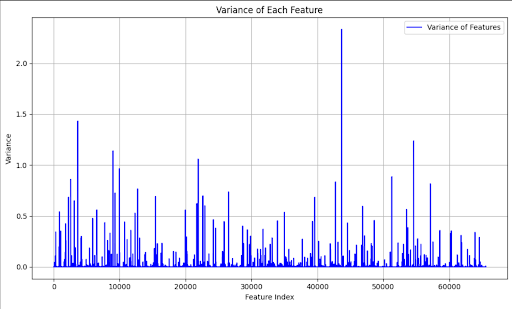}
        \caption{Activation variance for each feature}
        \label{fig:variance}
    \end{subfigure}
    \hfill
    \begin{subfigure}[b]{0.32\linewidth}
        \includegraphics[width=\linewidth]{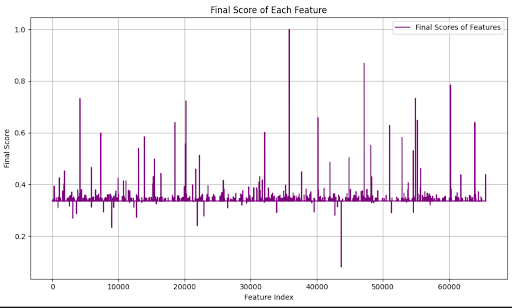}
        \caption{Final composite scores for all features}
        \label{fig:final_score}
    \end{subfigure}
    \caption{Feature activation analysis results. (a) Distribution of normalized activation differences showing outliers with strong differential responses. (b) Variance distribution revealing consistent vs. unreliable features. (c) Composite scores showing long-tailed distribution with few high-scoring candidates.}
    \label{fig:feature_analysis}
\end{figure}

Analyzing all of the 65,536 features in this layer showed distinct activation patterns when tested on the contrasting prompt pairs. Figure~\ref{fig:norm_diff} shows the normalized difference of the distribution across all of the features showing the base magnitude difference between an activation between harmful and harmless prompts.

Figure~\ref{fig:variance} shows the variance results from each feature activation pattern across the 100 contrasting prompt pairs. The variance distribution reveals that most of the features maintain a relatively constant activation, with low variance scores also clustered near zero.
Figure~\ref{fig:final_score} indicates the most valuable metric, composing the first 2 metrics using our scoring equation presented are the final composite scores. The distribution demonstrates a long-tail pattern as well, with a vast majority of features receiving a lower composite and only a small percentage achieving high scores above 0.5.

\subsubsection{Top-Performing Features Identification}

Table~\ref{tab:top_features} shows the eight highest-scored features from the composite score analysis. Feature 35831 achieved the maximum total composite score of 1.0, showing both the largest positive differential activation and highest consistency across prompt pairs. The rest of the features show a hierarchical distribution with feature 47156 scoring 0.869 and Feature 60211 achieving 0.785.

\begin{table}[ht]
  \caption{Top 8 highest feature scores out of all 65,536 features in the LLaMA 3 8B SAE release}
  \label{tab:top_features}
  \centering
  \begin{tabular}{p{2cm}p{3cm}p{4cm}}
    \toprule
    \textbf{Index} & \textbf{Feature Score} & \textbf{Normalized Diff. Sign} \\
    \midrule
    35831 & 1.000 & Positive \\
    47156 & 0.869 & Positive \\
    9000  & 0.799 & Negative \\
    60211 & 0.785 & Positive \\
    54916 & 0.733 & Positive \\
    20225 & 0.723 & Positive \\
    40185 & 0.658 & Positive \\
    55211 & 0.648 & Positive \\
    \bottomrule
  \end{tabular}
\end{table}

\subsection{Steering Performance Evaluation}

\subsubsection{Experimental Design and Feature Selection Strategy}

Our systematic evaluation tested four features representing distinct categories based on their differential activation patterns, enabling comprehensive assessment of our dual-strategy approach.

\textbf{Harmful-Activating Features (Positive norm\_diff\_mean):}
Feature 35831 was selected as our primary test case, achieving the highest composite score of 1.000 with strongly positive normalized difference, indicating preferential activation on harmful prompts. We applied negative steering strengths $[-4.0, -2.0]$ to suppress this feature's influence. Feature 43692 provided a secondary test from the harmful-activating category, selected for its high composite score and positive normalized difference, allowing comparison within this strategy.

\textbf{Safe-Activating Features (Negative norm\_diff\_mean):}
Feature 9000 was chosen for its strongly negative normalized difference, indicating preferential activation on safe prompts. We applied positive steering strengths $[0.5, 2.0, 4.0]$ to amplify this feature's protective effects. This selection tests the hypothesis that enhancing safe-activating features improves overall model safety performance.

\textbf{Control Feature:}
Feature 20000 served as our experimental control, selected for minimal differential activation between prompt types (norm\_diff\_mean $\approx 0$). This baseline allows assessment of steering effects on features with no clear safety relevance.

This experimental design systematically tests two complementary hypotheses: that suppressing harmful-activating features and amplifying safe-activating features both contribute to improved safety performance, while allowing measurement of their respective impacts on model utility.

\subsubsection{Feature 9000 and 43692 Steering Results}

\begin{figure}[ht]
    \centering
    \begin{subfigure}[b]{0.49\linewidth}
        \includegraphics[width=\linewidth]{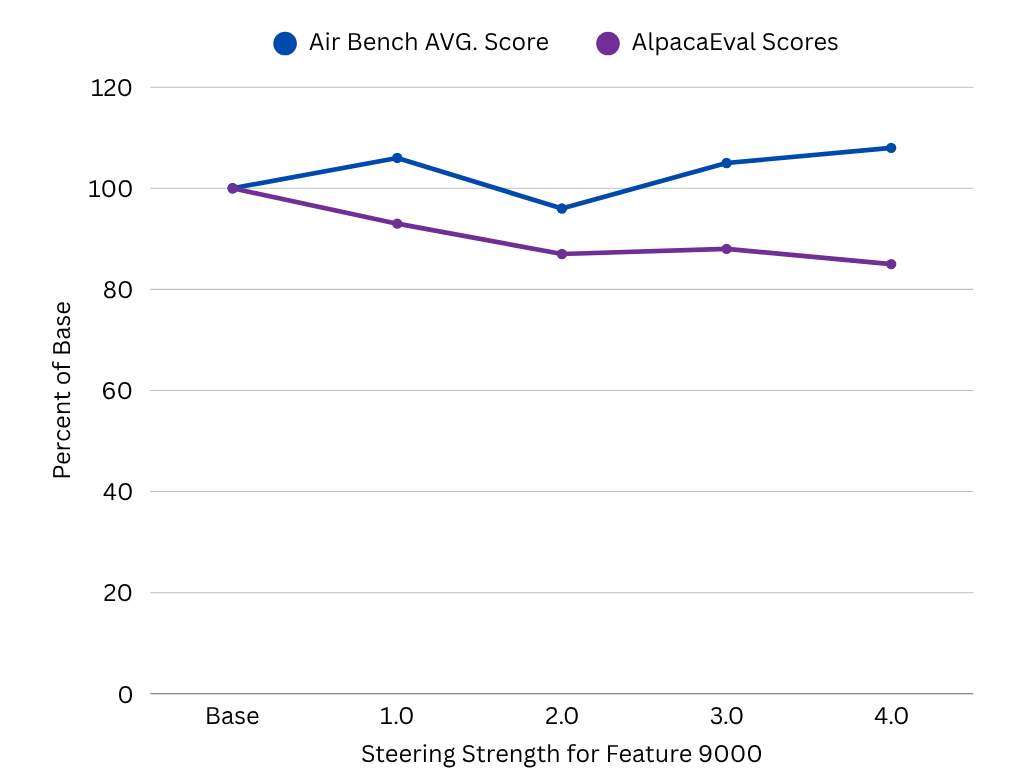}
        \caption{Feature 9000 Steering Results}
        \label{fig:feature9000}
    \end{subfigure}
    \hfill
    \begin{subfigure}[b]{0.49\linewidth}
        \includegraphics[width=\linewidth]{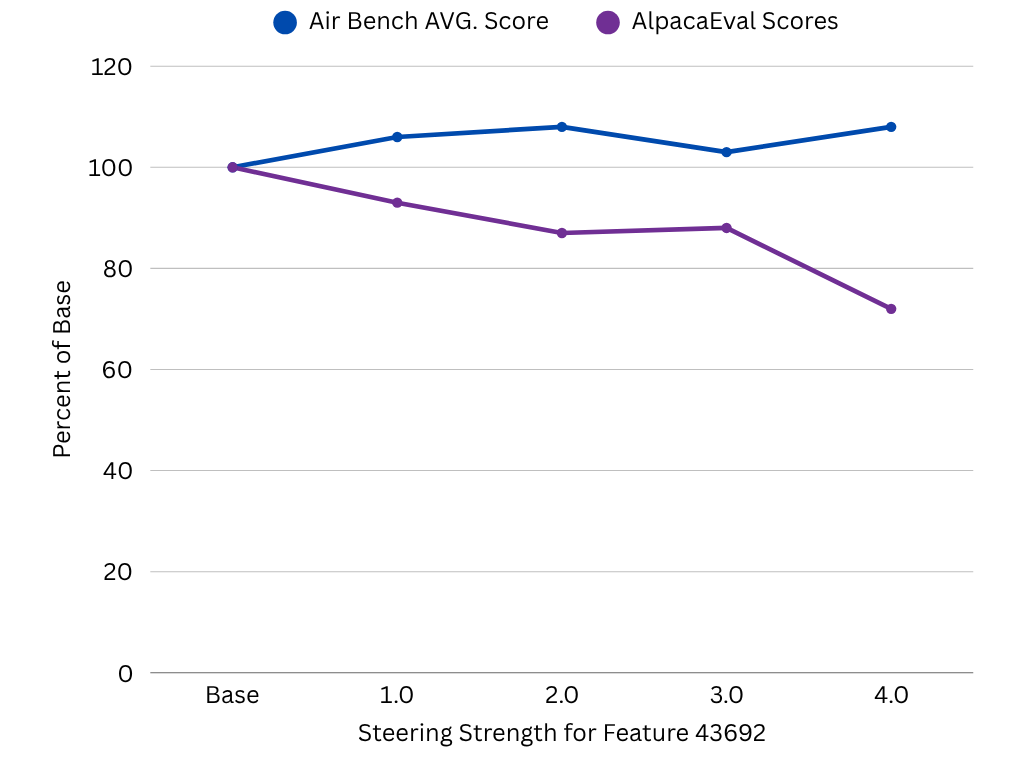}
        \caption{Feature 43692 Steering Results}
        \label{fig:feature43692}
    \end{subfigure}
    \caption{Steering results for features exhibiting a conventional safety-utility trade-off. (a) Steering Feature 9000 improves safety but degrades utility. (b) Steering Feature 43692 shows a similar pattern with a more severe utility drop at higher strengths.}
    \label{fig:tradeoff_features}
\end{figure}

Figure~\ref{fig:feature9000} demonstrates the results of steering on Feature 9000 across increasing steering strengths from the baseline to positive 4.0. Air Bench safety scores showed a modest improvement, with a peak of 108.8 at steering strength 4.0 representing an 8.8 percent increase in refusal detection from the baseline. AlpacaEval utility scores revealed steady degradation accompanying the safety improvements, declining from a baseline of 100 to 83.7 at steering strength 4.0, representing a 16.3 percent decrease in general model capability.

Figure~\ref{fig:feature43692} shows the characteristics of Feature 43692, implemented with negative steering to suppress its natural activation. Air Bench scores improved consistently, rising from 100 at baseline to 107.2 at strength 2.0 (7.4 percent improvement) and reaching 109.8 at maximum strength (10.0 percent improvement). However, AlpacaEval showed modest decline from 100 to 92.4 at steering strength 2.0 but fell to 74.1 at steering strength 4.0.

\subsubsection{Feature 35831 Steering Results}

\begin{figure}[ht]
    \centering
    \includegraphics[width=0.8\linewidth]{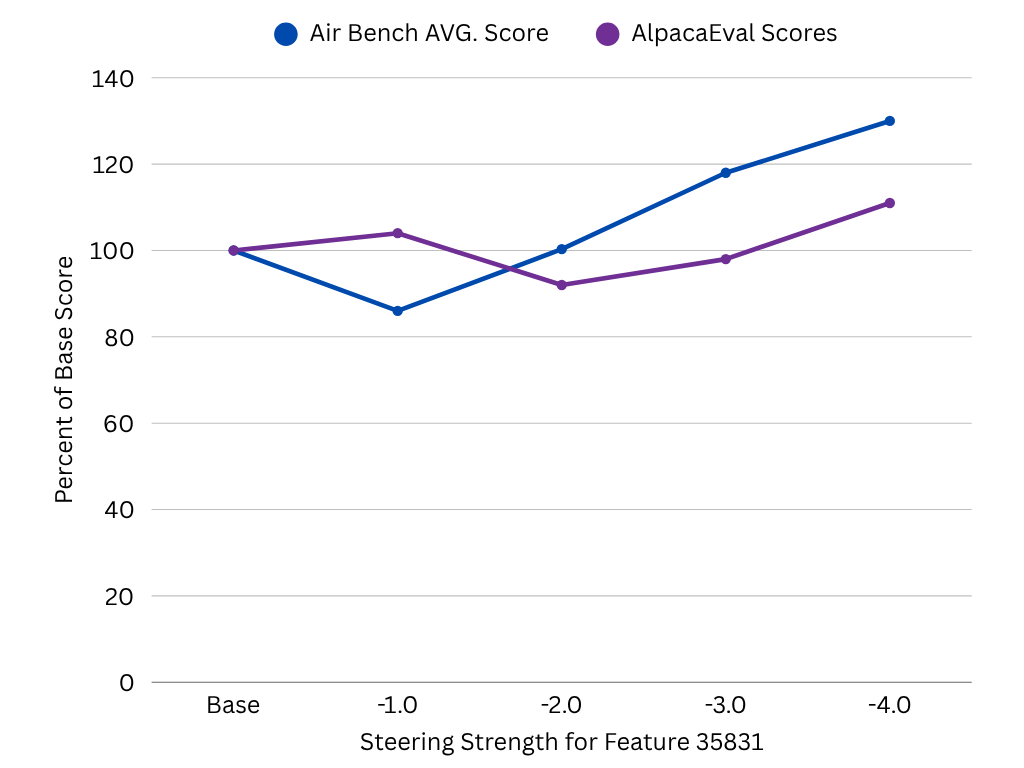}
    \caption{Feature 35831 Steering Results. This feature demonstrates simultaneous improvement in safety (AirBench score) and utility (AlpacaEval win rate), overcoming the typical trade-off.}
    \label{fig:feature35831}
\end{figure}

Figure~\ref{fig:feature35831} shows the performance of Feature 35831, the best performing feature according to the scoring system, also implemented with negative steering strength. Air Bench results showed substantial improvement from 100 to 118.9 at steering strength -2.0. Additionally, this safety improvement came with a utility boost, with AlpacaEval performance increasing from 100.0 to 111.1 at 4.0 steering strength.

\section{Discussion}

Our results show several key insights, the strong performance of Feature 35831 confirms our composite scoring methodology can identify features with real causal relationships to refusal behavior, moving beyond the heuristic approaches that characterize current literature \citep{marks_2024}. The effectiveness of SAEs in finding interpretable features within transformer models aligns with recent advances in mechanistic interpretability \citep{zhang_2025} and validates our systematic approach to feature selection.

\subsection*{Comparison with Traditional Approaches}
Traditional safety alignment methods like RLHF and Constitutional AI need extensive retraining and substantial computational resources \citep{ouyang_2022,bai_2022}. The SAE steering and composite score approach enables safety improvements using targeted specific features without requiring model retraining, addressing the computational efficiency concerns highlighted in recent work on scaling SAEs \citep{templeton_2024}. This approach can be applied to existing open-source models with immediate practical implications.
The ability to achieve both safety enhancement (18.9 percent improvement) and utility gains (11.1 percent improvement) shows a significant advantage over traditional methods, which normally require explicit safety-utility tradeoffs. This suggests that the SAE steering approach can unlock the model's capabilities by removing harmful patterns without constraining the model's behavior through additional training objectives.

\subsection*{Limitations and Methodological Considerations}
Our evaluation framework faces several important limitations affecting generalizability. The restriction to Llama-3 8B and Layer 25 limits understanding of scaling behaviors across architectures and transformer depths. While our contrasting prompt methodology provides systematic validation, broader domain coverage remains necessary.

\textbf{Computational Considerations:} While our approach avoids model retraining, SAE training represents substantial computational investment. However, pre-trained SAE weights can be reused across multiple steering applications, amortizing this cost. Our utilization of existing SAELens repository weights demonstrates practical deployment feasibility.

\textbf{Evaluation Robustness:} Our reliance on automatic judges (GPT-4o for AirBench, GPT-4 for AlpacaEval 2.0) introduces potential limitations despite demonstrated correlation with human preferences. Length bias effects and evaluation consistency represent areas requiring additional validation.

\textbf{Baseline Comparisons:} The absence of direct comparisons with alternative steering methods limits our ability to establish relative effectiveness claims. Resource constraints prevented systematic comparison, representing an important direction for future validation.

\section{Conclusion}

This work demonstrates that feature-guided SAE steering is a viable and efficient approach to improving the safety of LLMs without sacrificing utility, directly addressing current limitations in systematic feature selection and principled evaluation of safety-utility tradeoffs in SAE-based approaches. Our contributions include a novel contrasting prompt scoring method that systematically identifies safety-relevant features, moving beyond heuristic selection methods \citep{marks_2024}, paired with empirical validation that the method reliably predicts steering effectiveness.
The achievement of 18.9 percent safety and 11.1 percent utility enhancement with Feature 35831 represents a significant advance over traditional safety alignment approaches and demonstrates that principled SAE steering can unlock latent model capabilities while removing harmful interference patterns. This finding directly addresses the challenge that validating feature dictionaries in realistic scenarios without ground-truth remains elusive \citep{huang_2024} by providing systematic validation through comprehensive benchmarking.
The findings have immediate practical applications for LLM deployment, offering a computationally efficient alternative to traditional safety methods that require extensive retraining. While limitations need to be addressed to fully generalize the solution across different model architectures and scales, consistent with recent work on scaling SAEs \citep{templeton_2024}, the fundamental approach provides a solid foundation for future research in mechanistically-informed safety alignment.

\begin{ack}
We thank the reviewers for their valuable feedback and suggestions. This work was supported by computational resources provided by the academic institution.
\end{ack}

\bibliographystyle{plainnat}
\bibliography{references}


\appendix

\section{Neuronpedia Dashboard Example}
\label{sec:appendix_neuronpedia}

Although the quantitative scores from our contrasting prompt analysis were the primary driver for feature selection, we also used Neuronpedia's dashboard for qualitative validation and to gain deeper insight into feature behavior. For features available on the dashboard, it provides an auto-generated description, a list of top activating tokens, and visualizations of logit weights, which can help in hypothesis generation.

\begin{figure}[h]
    \centering
    \includegraphics[width=0.7\linewidth]{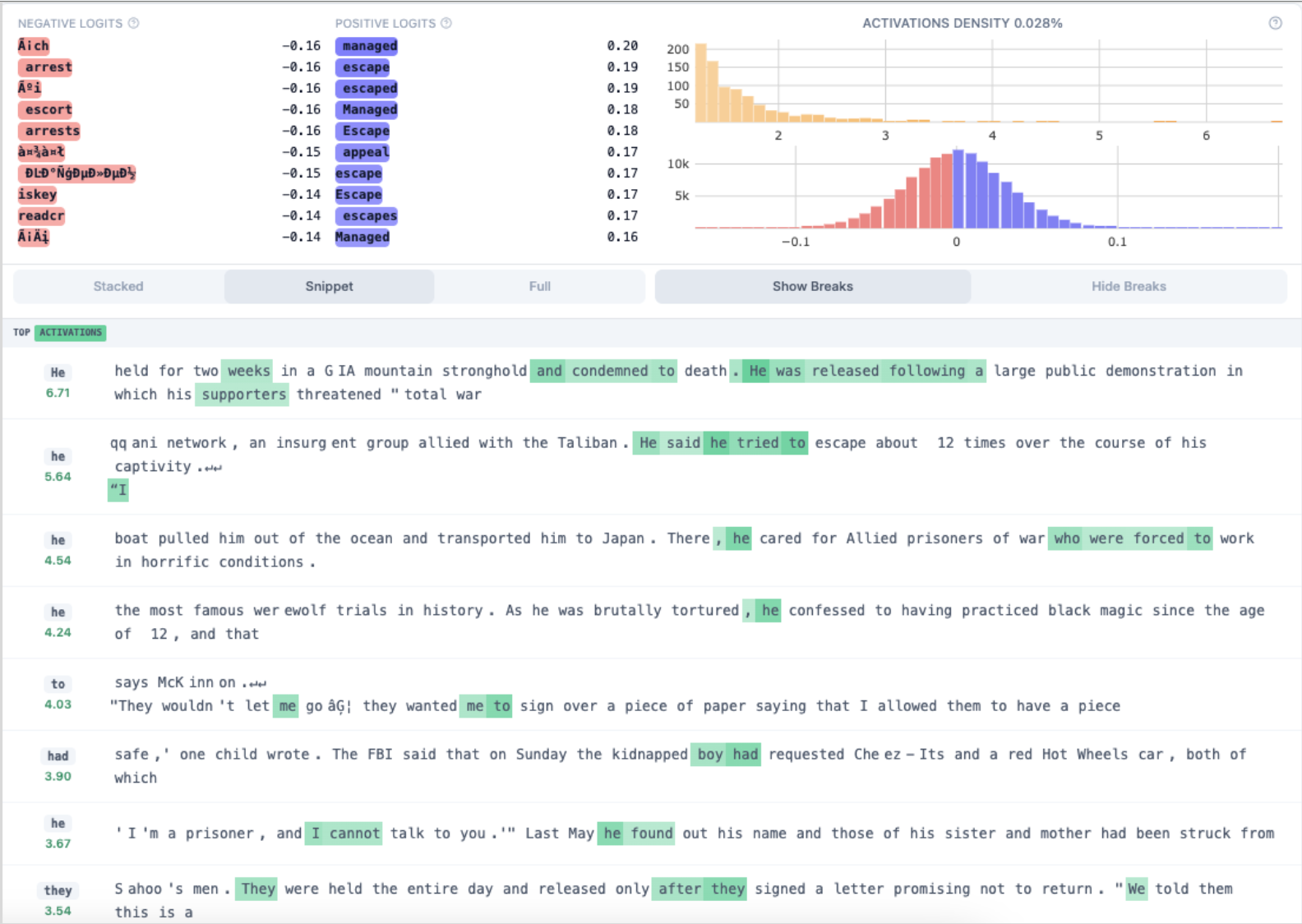}
    \caption{The Neuronpedia dashboard for Feature 1 in Llama 3 8B. This tool provides qualitative interpretations of a feature's function.}
    \label{fig:neuronpedia}
\end{figure}

As an illustrative example of the dashboard's interface, Figure~\ref{fig:neuronpedia} shows the analysis for Feature 1. While not a top-performing feature for our safety-steering task, it demonstrates the tool's capability to provide qualitative insights into a feature's function by summarizing its top activating tokens and logit weights. For features not already documented, a similar analysis could be generated using GPT-4.

\end{document}